\def\year{2020}\relax
\documentclass[letterpaper]{article} % DO NOT CHANGE THIS
\usepackage{aaai20}  % DO NOT CHANGE THIS
\usepackage{times}  % DO NOT CHANGE THIS
\usepackage{helvet} % DO NOT CHANGE THIS
\usepackage{courier}  % DO NOT CHANGE THIS
\usepackage[hyphens]{url}  % DO NOT CHANGE THIS
\usepackage{graphicx} % DO NOT CHANGE THIS
\usepackage{graphicx}  % DO NOT CHANGE THIS
\usepackage{amsfonts}
\usepackage{amsmath}
\usepackage{bm} % added by RS for bold vectors
\newcommand{\vect}[1]{\boldsymbol{\mathbf{#1}}}
\usepackage{changes}
\definechangesauthor[name={Sami}, color=red]{sk}
\definechangesauthor[name={Ruslan}, color=orange]{rs}
\usepackage{braket}
\usepackage{bm} % added by RS for bold vectors
\newcommand{\HsubM}{H_{\!M}}
\newcommand{\HsubC}{H_{\!C}}

\DeclareMathOperator*{\argmax}{arg\,max}

\usepackage[caption=false,font=footnotesize]{subfig}
\usepackage{float}
\usepackage{stfloats}

\title{Learning to Optimize Variational Quantum Circuits \\to Solve Combinatorial Problems}
\author{Sami Khairy,\textsuperscript{\rm 1} Ruslan Shaydulin,\textsuperscript{\rm 2} Lukasz Cincio,\textsuperscript{\rm 3} Yuri Alexeev,\textsuperscript{\rm 4} Prasanna Balaprakash\textsuperscript{\rm 4} \\
\textsuperscript{\rm 1}Illinois Institute of Technology, \textsuperscript{\rm 2}Clemson University, \textsuperscript{\rm 3}Los Alamos National Laboratory, \textsuperscript{\rm 4}Argonne National Laboratory\\skhairy@hawk.iit.edu, rshaydu@g.clemson.edu, lcincio@lanl.gov, yuri@anl.gov, pbalapra@anl.gov}

\begin{document}

\maketitle

\begin{abstract}
Quantum computing is a computational paradigm with the potential to outperform classical methods for a variety of problems. Proposed recently, the Quantum Approximate Optimization Algorithm (QAOA) is considered as one of the leading candidates for demonstrating quantum advantage in the near term. QAOA is a variational hybrid quantum-classical algorithm for approximately solving combinatorial optimization problems. The quality of the solution obtained by QAOA for a given problem instance depends on the performance of the classical optimizer used to optimize the variational parameters. In this paper, we formulate the problem of finding optimal QAOA parameters as a learning task in which the knowledge gained from solving training instances can be leveraged to find high-quality solutions for unseen test instances. To this end, we develop two machine-learning-based approaches. Our first approach adopts a reinforcement learning (RL) framework to learn a policy network to optimize QAOA circuits. Our second approach adopts a kernel density estimation (KDE) technique to learn a generative model of optimal QAOA parameters. In both approaches, the training procedure is performed on small-sized problem instances that can be simulated on a classical computer; yet the learned RL policy and the generative model can be used to efficiently solve larger problems. Extensive simulations using the IBM Qiskit Aer quantum circuit simulator demonstrate that our proposed RL- and KDE-based approaches reduce the optimality gap by factors up to $30.15$ when compared with other commonly used off-the-shelf optimizers. 
\end{abstract}

\section{Introduction}

Recently, a number of quantum computing devices have become available on the cloud. Current devices, which are commonly referred to as Noisy Intermediate-Scale Quantum (NISQ) devices, operate on a small number of qubits and have limited error-correction capabilities. Demonstrating quantum advantage on these devices, which is the ability to solve a problem more efficiently by using quantum computational methods compared with classical state-of-the-art methods, requires the development of algorithms that can run using a modest quantum circuit depth.

The Quantum Approximate Optimization Algorithm (QAOA) is one of the leading candidate algorithms for achieving quantum advantage in the near term. QAOA is a hybrid quantum-classical algorithm for approximately solving combinatorial problems \cite{farhi2014quantum}. QAOA combines a parameterized quantum state evolution that is performed on a NISQ device, with a classical optimizer that is used to find optimal parameters. %As other similar variational quantum-classical algorithm, such as Variational Quantum Eigensolver (VQE)~\cite{peruzzo2014variational} or Variational Quantum Classifier~\cite{Havlek2019,Schuld2019},
Conceivably, the quality of the solution produced by QAOA for a given combinatorial instance depends on the quality of the variational parameters found by the classical optimizer. Designing robust optimization methods for QAOA is therefore a prerequisite for achieving a practical quantum advantage.

Optimizing QAOA parameters is known to be a hard problem because the optimization objective is nonconvex with low-quality nondegenerate local optima %, and its gradients vanish as the dimension of variational parameters increase
\cite{shaydulin2019multistart,zhou2018quantum}. Many approaches have been applied to QAOA parameter optimization, including gradient-based \cite{romero2018strategies,zhou2018quantum,crooks2018performance} 
and derivative-free methods \cite{wecker2016training,yang2017optimizing,shaydulin2019multistart}.
Noting that the optimization objective of QAOA is specific to the underlying combinatorial instance, existing works approach the task of finding optimal QAOA parameters for a given instance as an exclusive task, and they devise methods that require quantum circuit evaluations on the order of thousands. To the best of our knowledge, approaching QAOA parameter optimization as a learning task is underexplored. To that end, we propose two machine-learning-based methods for QAOA parameter optimization, in which the knowledge gained from solving training instances can be leveraged to efficiently find high-quality solutions for unseen test instances with only a couple of hundred  quantum circuit evaluations. Our novel treatment of the QAOA optimization task has the potential to make QAOA a cost-effective algorithm to run on near-term quantum computers.  

The main contributions of our work are summarized as follows. First, we formulate the task of learning a QAOA parameter optimization policy as a reinforcement learning (RL) task. This approach can learn a policy network that can exploit geometrical regularities in the QAOA optimization objective of training instances, to efficiently optimize new QAOA circuits of unseen test instances. Second, we propose a sampling-based QAOA optimization strategy based on a kernel density estimation (KDE) technique.
This approach can learn a generative model of optimal QAOA parameters, which can be used to generate new parameters and quickly solve test instances. In both approaches, we choose a training set of small-sized combinatorial instances that can be simulated on a classical computer, yet the test set includes larger instances. We conduct extensive simulations using the IBM Qiskit Aer quantum circuit simulator to evaluate the performance of our proposed approaches. We show that the two approaches can reduce the optimality gap by factors up to $30.15$ when compared with other commonly used off-the-shelf optimizers. 

%\PB{consider removing it if you run out of space}
%The remainder of this paper is organized as follows. The QAOA algorithm is introduced in Section \ref{sec:QAOA}. Literature survey of related research work is given in Section \ref{sec:lit}. Our proposed machine-learning-based approaches for finding optimal QAOA parameters are presented in Section \ref{sec:ml}, followed by our results and discussion in Section \ref{seq:rs}. Finally, our concluding remarks and future work are given in Section \ref{sec:con}.

\section{The Quantum Approximate Optimization Algorithm}
\label{sec:QAOA}
The idea of encoding the solution to a combinatorial optimization problem in the spectrum of a quantum Hamiltonian goes back to 1989~\cite{Apolloni1989}. Using this quantum encoding, one can find the optimal solution to the original combinatorial problem by preparing the highest energy eigenstate of the problem Hamiltonian. Multiple approaches inspired by the adiabatic theorem~\cite{Kato1950} have been proposed to achieve this. Of particular interest is the Quantum Approximate Optimization Algorithm, introduced by~\cite{farhi2014quantum}, and its generalization, the Quantum Alternating Operator Ansatz~\cite{hadfield2017quantum}. QAOA is arguably one of the strongest candidates for demonstrating quantum advantage over classical approaches in the current NISQ computers era \cite{streif2019comparison}. 

In QAOA, a classical binary assignment combinatorial problem is first encoded in a cost Hamiltonian $H_C$ by mapping classical binary decision variables $s_i\in \{-1,1\}$ onto the eigenvalues of the quantum Pauli Z operator $\hat{\sigma}^z$. For unconstrained combinatorial problems, the initial state is a uniform superposition quantum state $|\psi \rangle = |+\rangle^{\otimes n}$, prepared by applying Hadamard gates on all qubits in the system. 
QAOA prepares a variational quantum state $\ket{\psi(\vect{\beta},\vect{\gamma})}$ by applying a series of alternating operators $e^{-i\beta_k H_M}$ and $e^{-i\gamma_k H_C}$, $\forall k \in [p]$, 

\begin{equation}
    \ket{\psi{(\vect{\beta},\vect{\gamma})}} =  e^{-i\beta_p \HsubM}e^{-i\gamma_p \HsubC}\cdots e^{-i\beta_1 \HsubM}e^{-i\gamma_1 \HsubC}\ket{+}^{\otimes n},
\label{eq:ansatz}
\end{equation}

\noindent where $\vect{\beta},\vect{\gamma} \in [-\pi, \pi]$ are $2p$ variational parameters, $n$ is the number of qubits or binary variables, and $H_M$ is the transverse field mixer Hamiltonian $H_M = \sum_i \hat{\sigma}_i^x$. Alternative initial states and mixer operators can be used to restrict $\ket{\psi{(\vect{\beta},\vect{\gamma})}}$ to the subspace of feasible solutions for constrained optimization problems~\cite{hadfield2017quantum}. In order to find the highest energy eigenstate of $H_C$, a classical optimizer is used to vary parameters $\vect{\beta},\vect{\gamma}$ to maximize the expected energy of $H_C$, 

\begin{equation}
    f(\vect{\beta},\vect{\gamma}) = \bra{\psi{(\vect{\beta},\vect{\gamma})}}\HsubC\ket{\psi{(\vect{\beta},\vect{\gamma})}}.
    \label{eq:obj}
\end{equation}

For $p\rightarrow\infty$, $\exists \vect{\beta}_*,\vect{\gamma}_* = \argmax_{\vect{\beta}, \vect{\gamma}}f(\vect{\beta},\vect{\gamma})$ such that the resulting quantum state $\ket{\psi{(\vect{\beta}_*,\vect{\gamma}_*)}}$ encodes the optimal solution to the classical combinatorial problem \cite{farhi2014quantum}. %However, QAOA is generally explored in the low-depth regime ($1\leq p\leq 10$) due to the limitations of near-term quantum hardware. 
In practice, the value $p$ is chosen based on the trade-off between the achieved approximation ratio, the complexity of parameter optimization, and the accumulated errors. Ideally, increasing $p$ monotonically improves the QAOA approximation ratio~\cite{zhou2018quantum}, although for higher $p$ the complexity of QAOA parameter optimization can limit the benefits~\cite{huang2019alibaba}. On real near-term quantum hardware, errors become a dominating factor. For instance, on a state-of-the-art trapped-ion quantum computer, increasing $p$ beyond $1$ does not lead to improvements in the approximation ratio because of errors~\cite{pagano2019quantum}. The hardware is rapidly progressing, however, and it is expected that QAOA with $1\leq p\leq 5$ can be run in the foreseeable future, thus motivating us to choose $p\in \{1, 2, 4\}$ for our benchmark. 

QAOA has been applied to a variety of problems, including graph maximum cut~\cite{crooks2018performance,zhou2018quantum}, network community detection~\cite{shaydulin2018network,shaydulin2018community}, and portfolio optimization, among many others~\cite{barkoutsos2019improving}.

\subsection{QAOA for Max-Cut}
In this paper, we explore QAOA applied to the graph maximum cut problem (Max-Cut). It is among the most commonly used target problems for QAOA because of its equivalence to quadratic unconstrained binary optimization. Consider a graph $G=(V,E)$ where $V$ is the set of vertices and $E$ is the set of edges. The goal of Max-Cut is to partition the set of vertices $V$ into two disjoint subsets such that the total weight of edges connecting the two subsets is maximized. Let binary variables $s_k$ denote the partition assignment of vertex $k$, $\forall k \in [n]$. Then Max-Cut can be formulated as follows,

\begin{equation} \label{maxcut2}
    \max_{\vect{s}}\sum_{i,j\in V} w_{ij} s_i s_j + c, \qquad s_k\in \{-1,1\}, \forall k
\end{equation}

\noindent where $w_{ij} = 1$ if $(i,j)\in E$, and $0$ otherwise, and $c$ is a constant. The objective in \eqref{maxcut2} can be encoded in a problem Hamiltonian by mapping binary variables $s_k$ onto the eigenvalues of the Pauli Z operator $\hat{\sigma}^z$: $H_C = \sum_{i,j\in V} w_{ij} \hat{\sigma}^z_i \hat{\sigma}^z_j$. The optimal binary assignment $\vect{s}$ for \eqref{maxcut2} is therefore encoded in the highest energy eigenstate of $H_C$. 
% \begin{equation} \label{ham}
%     H_C = \sum_{i,j\in V} w_{ij} \hat{\sigma}^z_i \hat{\sigma}^z_j,
% \end{equation}
% and the optimal binary assignment $\vect{s}$ for \eqref{maxcut2} is therefore encoded in the highest energy eigenstate of \eqref{ham}. 
Note that in the definition of $H_C$, there is an implicit tensor product with the identity unitary, $\hat{\sigma}^I$, applied to all qubits except for qubits $i,j$. 

In \cite{crooks2018performance} the author 
 shows that QAOA for Max-Cut can achieve approximation ratios exceeding those achieved by the classical Goemans-Williamson \cite{goemans1995improved} algorithm. A number of theoretical results show that QAOA for Max-Cut can improve on best-known classical approximation algorithms for some graph classes  \cite{osti_1492737,PhysRevA.97.022304}.

\subsection{Linear Algebraic Interpretation of QAOA}
Here, we provide a short linear algebraic description of QAOA for readers who are not familiar with the quantum computational framework. An $n$-qubit quantum state is a superposition (i.e., a linear combination) of computational basis states that form an orthonormal basis set in a complex vector space $\mathbb{C}^{2^n}$, 
\begin{equation}
\begin{aligned}
    \ket{\phi} &= \alpha_0 \ket{0..000}+\alpha_1 \ket{0..001}+\cdots+\alpha_{2^{n-1}} \ket{1..111} \\
    &=[\alpha_0, \alpha_1, \cdots,\alpha_{2^n-1}]^T,
\end{aligned}
\end{equation}
where $|\alpha_i|^2, \forall i \in [2^n-1]$ is the probability that the quantum state is in $\ket{i}$, $\sum_{i=0}^{2^n-1} |\alpha_i|^2 =1$. Quantum gates are unitary linear transformations on quantum states. For instance, Pauli Z and Pauli X operators and the identity operator are 
\begin{equation}
\hat{\sigma}^z = \begin{bmatrix}
1 & 0 \\ 
 0 &  -1
\end{bmatrix}, ~~\hat{\sigma}^x = \begin{bmatrix}
0 & 1 \\ 
 1 &  0
\end{bmatrix}, ~~ \hat{\sigma}^I = \begin{bmatrix}
1 & 0 \\ 
 0 &  1
\end{bmatrix}.  
\end{equation}
Note that the eigenvectors of the Pauli Z operator are computational basis states $\ket{0}$ and $\ket{1}$ with eigenvalues $1$ and $-1$, respectively. Therefore, $e^{-i\beta_k H_M}$ and $e^{-i\gamma_k H_C}$ are $2^n \times 2^n$ linear operators, which for large $n$ cannot be efficiently simulated classically. $H_C$ is a Hermitian linear operator with eigenvalues $\lambda_0, \lambda_1, \cdots, \lambda_{2^{n}-1}$. Based on the min-max variational theorem, the minimum eigenvalue $\lambda_\text{min}=\text{min}_{\vect{x}}~\{R_{H_C}(\vect{x}):\vect{x}\neq 0\}$, and the maximum eigenvalue $\lambda_\text{max}=\text{max}_{\vect{x}}~\{R_{H_C}(\vect{x}):\vect{x}\neq 0\}$, where $R_{H_C}(\vect{x})$ is the Rayleigh-Ritz quotient, $R_{H_C}(\vect{x}) = \frac{\vect{x}^\dagger H_C \vect{x}}{\vect{x}^\dagger\vect{x}}$, $\vect{x} \in \mathbb{C}^{2^n} \backslash \{\mathbf{0}\}$. Note that for any quantum state $\langle \phi| \phi \rangle = \phi^\dagger \phi = 1$, the highest energy of $H_C$ in \eqref{eq:obj} is in fact $\lambda_\text{max}$. That said, QAOA constructs linear operators parameterized by the $2p$ parameters, $\vect{\beta},\vect{\gamma}$, whose net effect is to transform the uniform superposition state, $\ket{\psi}=\frac{1}{\sqrt{2^n}}[1,\cdots,1]^T$, into a unit-length eigenvector that corresponds to $\lambda_\text{max}$.\footnote{Eigenvectors of a quantum Hamiltonian $H_C$ are referred to as its eigenstates; therefore, ``transforming $\ket{\psi}$ into an eigenvector of $H_C$ corresponding to $\lambda_\text{max}$'' is a different way of saying ``preparing an eigenstate of $H_C$ corresponding to the energy $\lambda_\text{max}$.''}

\section{Related Works}
In this section, a literature review of related research work is presented.
\label{sec:lit}

\subsection{QAOA Parameter Optimization}
A number of approaches have been explored for QAOA parameter optimization, including a variety of off-the-shelf gradient-based~\cite{romero2018strategies,crooks2018performance} and derivative-free methods~\cite{wecker2016training,yang2017optimizing,shaydulin2019multistart}. QAOA parameters have been demonstrated to be hard to optimize by off-the-shelf methods~\cite{shaydulin2019multistart} because the energy landscape of QAOA is nonconvex with low-quality nondegenerate local optima~\cite{zhou2018quantum}. Off-the-shelf optimizers ignore features of QAOA energy landscapes and nonrandom concentration of optimal parameters. Researchers have demonstrated theoretically that for certain bounded-degree graphs~\cite{brandao2018fixed},  
QAOA energy landscapes are graph instance-independent given instances that come from a reasonable distribution.  Motivated by this knowledge, we develop two machine-learning-based methods exploiting the geometrical structure of QAOA landscapes and the concentration of optimal parameters, so that the cost of QAOA parameter optimization can be amortized.

\subsection{Hyperparameter and Optimizer Learning}
Hyperparameter optimization, which involves the optimization
of hyperparameters used to train a machine learning model, is an active field of research \cite{automl_book}. Each hyperparameter choice corresponds to one optimization task, and thus tuning of hyperparamters can be regarded as a search over different optimization instances, which is analogous to our problem formulation. Recent methods devise sequential model-based Bayesian optimization techniques \cite{feurer2015initializing} or asynchronous parallel model-based search \cite{8638041}, while older methods devise random-sampling-based strategies \cite{prasanna2007}.

On the other hand, learning an optimizer to train machine learning models has recently attracted considerable research interests. The motivation is to design optimization algorithms that can exploit structure within a class of problems, which is otherwise unexploited by hand-engineered off-the-shelf optimizers. In existing works, the learned optimizer is implemented by long short-term memory \cite{andrychowicz2016learning,verdon2019learning} or a policy network of an RL agent \cite{li2016learning}. Our RL-based approach to optimizer learning differs from that of \cite{li2016learning} mainly in the choice of reward function and the policy search mechanism. In our work, a Markovian reward function is chosen to improve the learning process of a QAOA optimization policy. 

\section{Learning Optimal QAOA Parameters}
\label{sec:ml}

Since evaluating a quantum circuit is an expensive task, the ability to find good variational parameters using a small number of calls to the quantum computer is crucial for the success of QAOA. To amortize the cost of QAOA parameter optimization across graph instances, we formulate the problem of finding optimal QAOA parameters as a learning task. We choose a set of graph instances, $G_\text{Train}$, which include graph instances that are representative of their respective populations, as training instances for our proposed machine learning methods. The learned models can then be used to find high-quality solutions for unseen instances from a test set $G_\text{Test}$. In Section \ref{RLsol}, we propose an RL-based approach to learn a QAOA parameter optimization policy. In Section \ref{KDEsol}, we propose a KDE-based approach  to learn a generative model of optimal QAOA parameters. Section \ref{Data} elaborates on graph instances in $G_\text{Train}$ and $G_\text{Test}$.

\subsection{Learning to Optimize QAOA Parameters with Deep Reinforcement Learning} \label{RLsol}

Our first approach aims to learn a QAOA parameter optimization policy, which can exploit structure and geometrical regularities in QAOA energy landscapes to find high-quality solutions within a small number of quantum circuit evaluations. This approach can potentially outperform hand-engineered that are designed to function in a general setting. We cast the problem of learning a QAOA optimizer as an RL task, where the learned RL policy is used to produce iterative parameter updates, in a way that is analogous to hand-engineered iterative optimizers.  

In the RL framework, an autonomous agent learns how to map its state in a state space, $s \in \mathcal{S}$, to an action from its action space, $a \in \mathcal{A}$, by repeated interaction with an environment. The environment provides the agent with a reward signal, $r \in \mathbb{R}$, in response to its action. Based on the reward signal, the agent either reinforces the action or avoids it at future encounters, in an attempt to maximize the expected total discounted rewards received over time \cite{sutton2018reinforcement}. 

Mathematically, the RL problem is formalized as a Markov decision process (MDP), defined by a tuple $(\mathcal{S},\mathcal{A},\mathcal{P}, \mathcal{P}_0,\mathcal{R},\zeta)$, where the environment dynamics, $\mathcal{P}:\mathcal{S}\times \mathcal{A} \times \mathcal{S} \rightarrow [0,1]$, that is,  the model's state-action-state transition probabilities, $\mathbf{P}(s_{t+1}|s_t,a_t)$, and $\mathcal{P}_0:\mathcal{S} \rightarrow [0,1]$, which is the initial distribution over the states, are unknown; $\mathcal{R}: \mathcal{S} \times \mathcal{A} \times \mathcal{S} \rightarrow \mathbb{R}$ is a reward function that guides the agent through the learning process; and $\zeta$ is a discount factor to bound the cumulative rewards and trade off how farsighted or shortsighted the agent is in its decision making. A solution to the RL problem is a stationary Markov policy that maps the agent's states to actions, $\pi(a|s)$, such that the expected total discounted rewards is maximized, 

\begin{equation} \label{obj2}
\mathbf{\pi^*} = \arg\max_{\pi} ~~~\mathbb{E}[\sum_{t=0}^T \zeta^t \mathcal{R}(s_t, a_t, s_{t+1})].
\end{equation}

In our setting, we seek a Markov policy that can be used to produce iterative QAOA parameter updates. This policy must exploit QAOA structure and geometrical regularities such that high-quality solutions can be achieved despite the challenges pertaining to QAOA parameter optimization. To do so, we formulate QAOA optimizer learning as an MDP as follows. 
\begin{enumerate}
\item  $\forall s_t \in \mathcal{S}$, $s_t = \{\Delta f_{tl}, \Delta \vect{\beta}_{tl},\Delta \vect{\gamma}_{tl} \}_{l=t-1,...,t-L}$; that is, the state space is the set of finite differences in the QAOA objective and the variational parameters between the current iteration and $L$ history iterations, $\mathcal{S} \subset \mathbb{R}^{(2p+1)L}$.
\item $\forall a_t \in \mathcal{A}$, $a_t = \{\Delta \vect{\beta}_{tl}, \Delta \vect{\gamma}_{tl} \}_{l=t-1}$; that is, the action space is the set of step vectors  used to update the variational parameters, $\mathcal{A} \subset \mathbb{R}^{2p}$.
\item $\mathcal{R}(s_t, a_t, s_{t+1}) = f(\vect{\beta}_t + \Delta \vect{\beta}_{t,tl}, \vect{\gamma}_t + \Delta \vect{\gamma}_{tl}) - f(\vect{\beta}_t, \vect{\gamma}_t), l=t-1$; that is, the reward is the change in the QAOA objective between the next iteration and the current iteration. 
\end{enumerate}

The motivation behind our state space formulation comes from the fact that parameter updates at $\vect{x}_t = (\vect{\beta}_{t},\vect{\gamma}_{t})$ should be in the direction of the gradient at $\vect{x}_t$ and the step size  should be proportional to the Hessian at $\vect{x}_t$, both of which can be numerically approximated by using the method of finite differences. The RL agent's learning task is therefore to find the optimal way of producing a step vector $a_t = \{\Delta \vect{\beta}_{tl}, \Delta \vect{\gamma}_{tl} \}_{l=t-1}$, given some collection of historical differences in the objective and parameters space, $\{\Delta f_{tl}, \Delta \vect{\beta}_{tl},\Delta \vect{\gamma}_{tl} \}_{l=t-1,...,t-L}$, such that the expected total discounted rewards are maximized. Note that \eqref{obj2} is maximized when $\mathcal{R}(s_t, a_t, s_{t+1}) \geq 0$, which means the QAOA objective has been increased between any two consecutive iterates. The choice of reward function adheres to the Markovian assumption and encourages the agent to take strides in the landscape of the parameter space that yield a higher increase in the QAOA objective \eqref{eq:obj}, if possible, while maintaining conditional independence on historical states and actions.

\subsubsection{Training Procedure} Learning a generalizable RL policy that can perform well on a wide range of test instances requires the development of a proper training procedure and reward normalization scheme. We use the following strategy to train  the RL agent on instances in $G_\text{Train}$. A training episode is defined to be a trajectory of length $T=64$, which is sampled from a depth-$p$ QAOA objective \eqref{eq:obj} corresponding to one of the training instances in $G_\text{Train}$. At the end of each episode, the trajectory is cut off and restarted from a random point in the domain of \eqref{eq:obj}. Training instances are circulated in a round-robin fashion, with each episode  mitigating destructive policy updates and overfitting. Rewards for a given training instance are normalized by the mean depth-$p$ QAOA objective corresponding to that instance, which is estimated a priori by uniformly sampling the $2p$ variational parameters. Training is performed for $750$ epochs of $128$ episodes each, and policy updates are performed at the end of each epoch. 

\subsubsection{Deep RL Implementation} 

We train our proposed deep RL framework using the actor-critic Proximal Policy Optimization (PPO) algorithm \cite{schulman2017proximal}. In PPO, a clipped surrogate advantage objective is used as the training objective,
\begin{equation} \label{ppo}
\resizebox{1\hsize}{!}{$
L^\text{clip}(\theta) = \hat{\mathbb{E}}_t[\text{min}( \frac{\pi_\theta (a_t|s_t)}{\pi_{\theta_\text{old}}(a_t|s_t)} \hat{A}_t, \text{clip}(\frac{\pi_\theta (a_t|s_t)}{\pi_{\theta_\text{old}}(a_t|s_t)}, 1+\epsilon, 1-\epsilon)\hat{A}_t) ]$}
\end{equation}
The surrogate advantage objective, $\hat{\mathbb{E}}_t[\frac{\pi_\theta (a_t|s_t)}{\pi_{\theta_\text{old}}(a_t|s_t)} \hat{A}_t]$, is a measure of how the new policy performs relative to the old policy. Maximizing the surrogate advantage objective without constraints could lead to large policy updates that may cause policy collapse. To mitigate this issue, PPO maximizes the minimum of an unclipped and a clipped version of the surrogate advantage, where the latter removes the incentive for moving  $\frac{\pi_\theta (a_t|s_t)}{\pi_{\theta_\text{old}}(a_t|s_t)}$ outside of  $[1-\epsilon, 1+\epsilon]$. Clipping therefore acts as a regularizer that controls how much the new policy can go away from the old one while still improving the training objective. In order to further ensure reasonable policy updates, a simple early stopping method is adopted, which terminates gradient optimization on \eqref{ppo} when the mean KL-divergence between the new and old policy hits a predefined threshold. 

Fully connected multilayer perceptron networks with two hidden layers for both the actor (policy) and critic (value) networks are used. Each hidden layer has $64$ neurons. $Tanh$ activation units are used in all neurons. The range of output neurons is scaled to $[-0.1,0.1]$. The discount factor and number of history iterations in the state formulation are set to $\zeta=0.99$ and $L=4$, respectively. %\PB{What is L where is it defined?}.
A Gaussian policy with a constant noise variance of $e^{-6}$ %\PB{$10^{-6}$?} exp(-6) is correct
is adopted throughout the training in order to maintain constant exploration and avoid getting trapped in a locally optimal policy. At testing, the trained policy network corresponding to the mean of the learned Gaussian policy is used, without noise.

\subsection{Learning to Sample Optimal QAOA Parameters with Kernel Density Estimation} \label{KDEsol}

In our second approach, we aim to learn the distribution of optimal QAOA parameters and use this distribution to sample QAOA parameters that can provide high-quality solutions for test instances. Although previous theoretical results show that optimal QAOA parameters concentrate for graph instances \textit{necessarily} coming from the same distribution \cite{brandao2018fixed}, we learn a meta-distribution of optimal QAOA parameters for graph instances in $G_\text{Train}$, which come from diverse classes and distributions. This approach can potentially eliminate the need for graph feature engineering and  similarity learning and can  dramatically simplify the solution methodology. 

To learn a meta-distribution of optimal QAOA parameters, we adopt a KDE technique. Suppose we have access to a set of $N$ optimal QAOA parameters $S_*^p = \{\vect{x}_*^i = (\vect{\beta}_{*}^i,\vect{\gamma}_{*}^i) \}_{i=0}^{N-1}$ for graph instances in $G_\text{Train}$ and a given QAOA circuit depth, $p$. A natural local estimate for the density $\hat{f}_{\vect{X}}$ at $\vect{x}$ is $\hat{f}_{\vect{X}}(\vect{x}) = \frac{\# \vect{X}^i_* \in \mathcal{B}(\vect{x})}{N\omega}$, where $\mathcal{B}(\vect{x})$ is a neighborhood of width $\omega$ around $\vect{x}$ based on some distance metric. This estimate, however, is not smooth. %\PB{do you mean nonsmooth and nonconvex}
Instead, one commonly  adopts a Parzen-Rosenblatt approach with a Gaussian kernel to obtain a smoother estimate, 

\begin{equation} \label{dest}
\hat{f}_{\vect{X}}(\vect{x}) = \frac{1}{N} \sum_{i=0}^{N-1} K_\omega (\vect{x},\vect{x}^i_*)
\end{equation}
where $K_\omega (\vect{x},\vect{x}^i_*) = \frac{1}{(2 \pi \omega^2)^{N/2}} \text{exp}(\frac{-(\vect{x}-\vect{x}^i_*)^T(\vect{x}-\vect{x}^i_*)}{2\omega^2})$. 

In order to generate new QAOA parameters $\vect{z}$ using \eqref{dest}, a data point $\vect{x}_*^i$ from $S_*^p$ is chosen uniformly randomly with probability $1/N$, and it is added to a sample $\vect{X}^\prime$ $\sim \mathcal{N}(\mathbf{0}, \omega^2 \mathbb{I})$, that is, a sample drawn from a multivariate Gaussian distribution with zero mean and diagonal covariance matrix $\omega^2 \mathbb{I}$. Noting that $\hat{f}_{\vect{X}^\prime}(\vect{x}) = K_\omega (\vect{x},\mathbf{0})$, one can easily  confirm that this sampling strategy yields data points $\vect{z}$ distributed according to \eqref{dest},

\begin{equation}
\begin{aligned}
\mathbb{P}(\vect{Z}\leq \vect{z}) &= \hat{F}_{\vect{Z}}(\vect{z}) = \mathbb{P}(\vect{X}+\vect{X}^\prime \leq \vect{z})\\
&= \sum_{i=0}^{N-1} \mathbb{P}(\vect{X}+\vect{X}^\prime \leq \vect{z} | \vect{X} = \vect{x}_*^i) \mathbb{P}(\vect{X}=\vect{x}_*^i)\\
&=\frac{1}{N}\sum_{i=0}^{N-1} \mathbb{P}(\vect{X}^\prime \leq \vect{z}-\vect{x}_*^i)\\
&=\frac{1}{N} \sum_{i=0}^{N-1}\hat{F}_{\vect{X}^\prime}(\vect{z}-\vect{x}_*^i)
\end{aligned}
\end{equation}
and hence, $\hat{f}_{\vect{Z}}(\vect{z}) =\frac{1}{N} \sum_{i=0}^{N-1}K_\omega(\vect{z}-\vect{x}_*^i,\mathbf{0})=\frac{1}{N} \sum_{i=0}^{N-1}K_\omega(\vect{z},\vect{x}_*^i)$. 

This methodology assumes that $S_*^p$ is readily available. In order to construct $S_*^p$ in practice, a derivative-free off-the-shelf optimizer is started from a large number ($\sim 10,000$ as in \cite{zhou2018quantum}) of random points in the domain of \eqref{eq:obj}. Because \eqref{eq:obj} is known to be a nonconvex function in the parameter space, a portion of these extensive searches may converge to low-quality local optima. To tackle this issue, we admit into $S_*^p$ only parameters that achieve an optimality ratio of $99\%$ or higher for a given training instance and a given QAOA circuit depth. 

\subsection{Graph Max-Cut Instances}
\label{Data}
\begin{figure}[htbp]
\centering
\includegraphics[width=0.79\columnwidth]{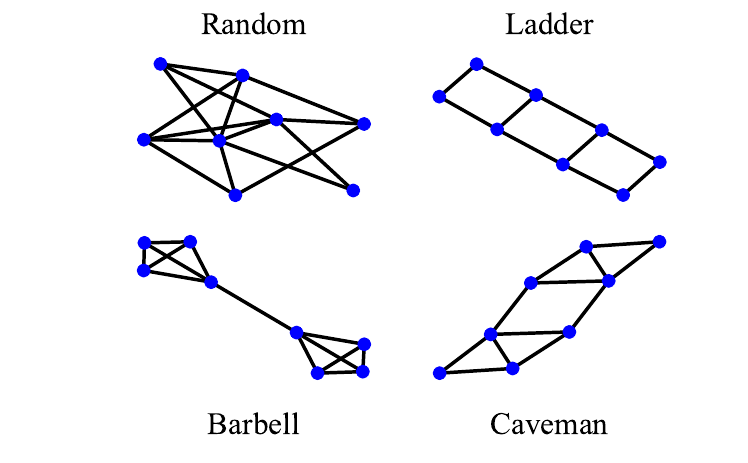}
\caption{Sample graph instances: random $G_R(n_R=8, e_p=0.5)$, ladder $ G_L(n_L=4)$, barbell $G_B(n_B=4)$, and caveman $G_C(n_C=2, n_k=4)$.}
\label{fig:GRAPHS}
\end{figure}

In this subsection, we describe graph instances in $G_\text{Train}$ and $G_\text{Test}$. Four classes of graphs are considered: (1) Erdos-Renyi random graphs $G_R(n_R, e_p)$, where $n_R$ is the number of vertices and $e_p$ is the edge generation probability, (2) ladder graphs $G_L(n_L)$, where $n_L$ is the length of the ladder, (3) barbell graphs $G_B(n_B)$, formed by connecting two complete graphs $K_{n_B}$ by an edge, and (4) Caveman graphs $G_C(n_C, n_k)$, where $n_C$ is the number of cliques and $n_k$ is the size of each clique. Figure \ref{fig:GRAPHS} shows sample graph instances drawn from each graph class. 

To construct $G_\text{Train}$, we choose one representative graph instance of $8$ vertices from each class and distribution, amounting to $|G_\text{Train}|=7$ training instances. $G_\text{Test}$ contains $94$ instances with varying number of vertices, as shown in Table \ref{table2}. We choose $|G_\text{Train}| < |G_\text{Test}|$ to demonstrate that combining our proposed machine learning approaches with QAOA can be a powerful tool for amortizing the QAOA optimization cost across graph instances. In addition, we choose to train on graph instances that are smaller than test instances, to demonstrate that our proposed approaches are independent of instance size or complexity. We  note that $G_\text{Train} \cap G_\text{Test} = \{ \}$. 

\begin{table}[t] 
\caption{Train and test graph Max-Cut instances.}\smallskip
\centering
\resizebox{.95\columnwidth}{!}{
\smallskip\begin{tabular}{|l|l|l|}
\hline
Graph Class & $G_\text{Train}$ & $G_\text{Test}$ \\
\hline \hline 
$G_R(n_R, e_p)$ & $n_R = 8$, & $n_R \in \{8,12,16,20 \}$, seed $=\{1,2,3,4\}$ \\
& $e_p \in \{0.5,0.6,0.7,0.8\}$ & $e_p \in \{0.5,0.6,0.7,0.8\}$ \\
\hline
$ G_L(n_L)$ & $n_L = 4$ & $n_L \in \{2,3,5,6,7,8,9,10,11 \}$ \\
\hline
$G_B(n_B)$ & $n_B = 4$ & $n_B \in \{3,5,6,7,8,9,10,11\}$ \\
\hline
$G_C(n_C, n_k)$ & $(n_C,n_K)=(2,4) $ &  $\{(n_C,4):n_C \in \{3,4,5\}\}$,\\
& & $\{(n_C,3):n_C \in \{3,5,7\}\}$,\\
& & $\{ (2,n_K): n_K \in \{3,5,6,7,8,9,10\} \}$\\
\hline \hline 
 & $|G_\text{Train}|=7$ & $|G_\text{Test}|$ = 94\\
\hline 
\end{tabular} 
}
\label{table2}
\end{table}
\section{Results and Discussion}
\label{seq:rs}

\begin{figure*}[ht]
\subfloat[$G_R(n_R=16, e_p=0.5)$]{\includegraphics[width=0.5\columnwidth]{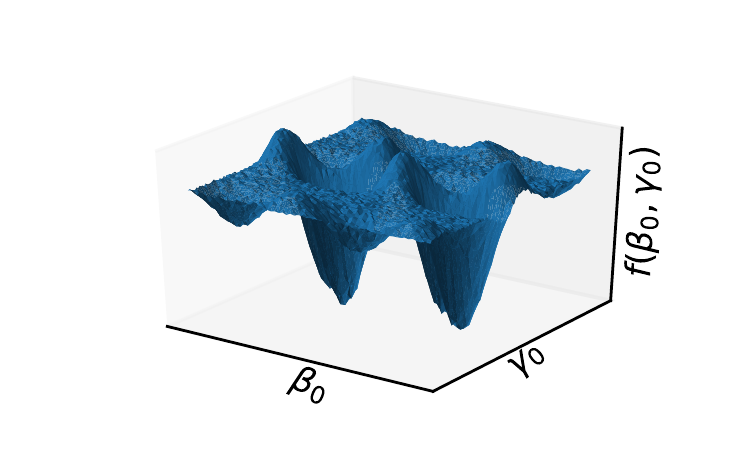}}
\hfill
\subfloat[$G_L(n_L=8)$]{\includegraphics[width=0.5\columnwidth]{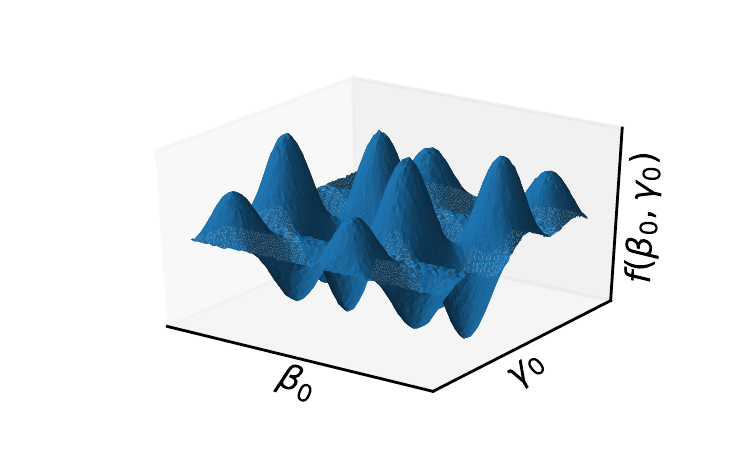}} 
\hfill
\subfloat[$G_B(n_B=8)$]{\includegraphics[width=0.5\columnwidth]{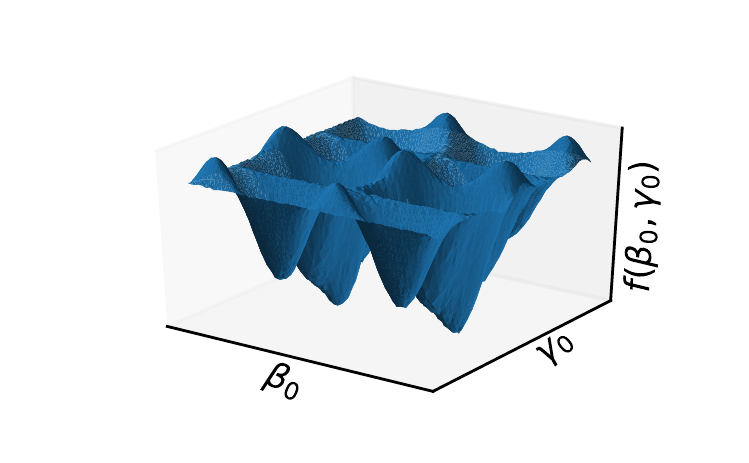}}
\hfill
\subfloat[$G_C(n_C=2, n_k=8)$]{\includegraphics[width=0.5\columnwidth]{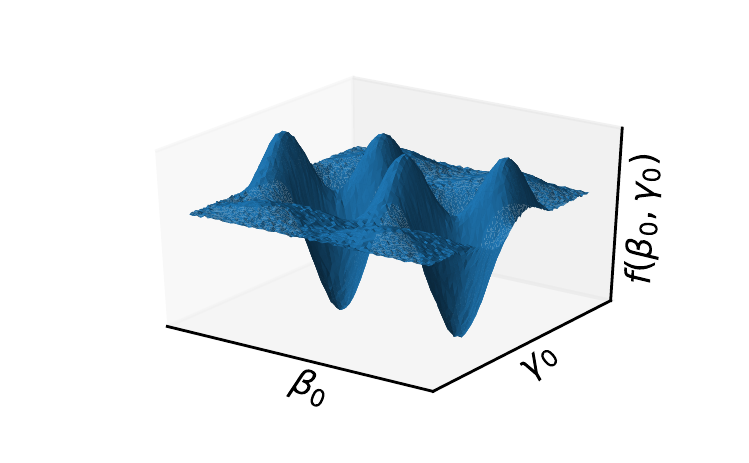}} 
\hfill
\caption{QAOA energy landscapes for $p=1$ (a) random graph, (b) ladder graph, (c) barbell graph, and (d) Ccveman graph.}
\label{fig:energyLandscapes}
\end{figure*}

In this section, we present the results  of our work. We use Qiskit Aer to perform noiseless simulations of QAOA circuits. % Noiseless is important!
In Figure \ref{fig:energyLandscapes} we show the expected energy landscape of the cost Hamiltonian for a depth $(p=1)$ QAOA circuit with two variational parameters $\beta_0$ and $\gamma_0$ for some graph instances. We can see that the expected energy of the cost Hamiltonian is nonconvex in the parameter space and is noisy because it is statistically estimated based on quantum circuit measurements. These features tend to be more severe as the depth of the QAOA circuit increases, thus posing serious challenges for commonly used derivative-free off-the-shelf optimizers.

\begin{figure*}[ht]
\centering
\subfloat[Learning curve of RL agent. At the beginning of training, RL performance on training instances is as good as a random-based sampling strategy. As training progresses, the learned RL policy collects higher expected discounted rewards. ]{\includegraphics[width=0.66\columnwidth]{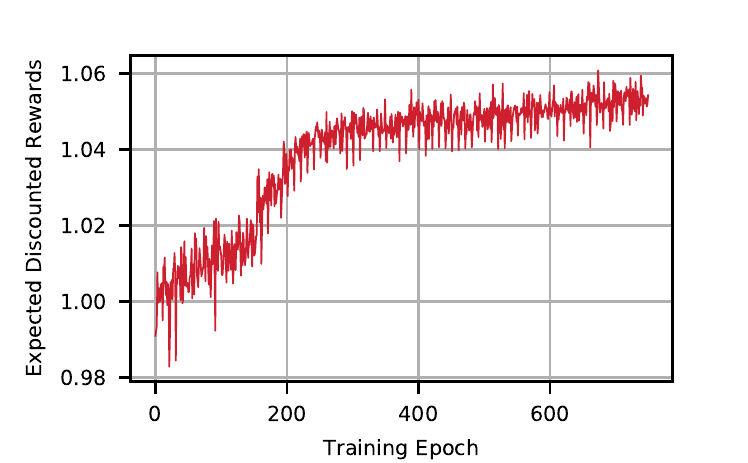}}
\hfill
\centering
\subfloat[Performance of RL agent on one training instance during training. The policy network at the end of training (epoch $740$) could produce a trajectory that reaches a high value quickly.]{\includegraphics[width=0.66\columnwidth]{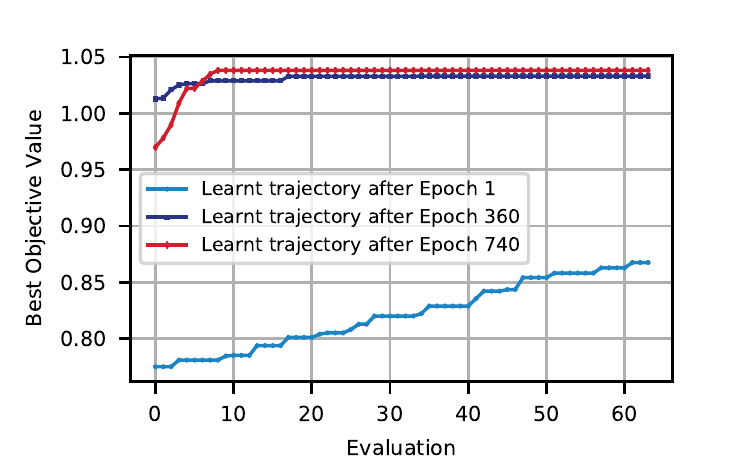}} 
\hfill
\centering
\subfloat[Optimal QAOA parameters density learned via KDE. Concentration of optimal parameters in some regions of the parameters space is evident.]{\includegraphics[width=0.66\columnwidth]{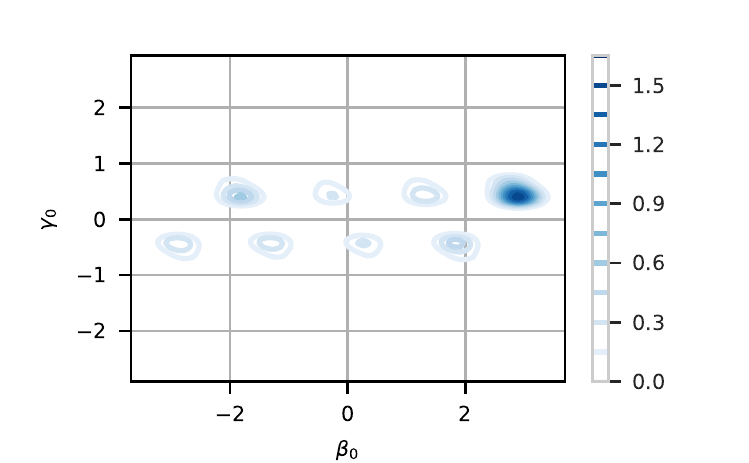}} 
\caption{RL and KDE training-related results ($p=1$)}
\label{fig:drl}
\end{figure*}

In Figure \ref{fig:drl}, we present training results associated with our proposed deep RL- and KDE-based approaches for a depth $(p=1)$ QAOA circuit. The RL learning curve during the training procedure is shown in Figure \ref{fig:drl}(a). We can see that the expected total discounted rewards of the RL agent starts around $1$ at the beginning of training, which means the RL agent is performing as well as random sampling. As training progresses, the performance of the learned optimization policy on the training set improves. This can also be seen in Figure \ref{fig:drl}(b), which shows the best objective value for one of the training instances versus time steps of an episode  at different stages of training. The optimization policy learned at the end of training, namely, epoch $750$, produces a trajectory that rises quickly to a higher value compared with the trajectories produced by the optimization policies learned at the middle and beginning of training. On the other hand, Figure \ref{fig:drl}(c) shows contour lines for the learned bivariate probability density based on $S_*^1$ using our proposed KDE technique. We can see that optimal QAOA parameters for a depth $(p=1)$ QAOA circuits of training instances in $G_\text{Train}$ concentrate in some regions in the domain of \eqref{eq:obj}.  

Next, we benchmark the performance of our trained RL-based QAOA optimization policy and the sampling-KDE-based QAOA optimization strategy by comparing their performance with common derivative-free off-the-shelf optimizers  implemented in the NLopt nonlinear optimization package~\cite{nlopt}, namely, BOBYQA, COBYLA, and Nelder-Mead, 
%namely BOBYQA~\cite{powell2009bobyqa}, COBYLA~\cite{powell1994direct,powell1998direct}, and Nelder-Mead~\cite{nelder1965simplex}, 
as well as a purely random sampling-based strategy. Starting from 10 randomly chosen variational parameters in the domain of \eqref{eq:obj}, each optimizer is given $10$ attempts with a budget of $B=192$ quantum circuit evaluations to solve QAOA energy landscapes corresponding to graph instances in $G_\text{Test}$. In each of the $10$ attempts, the random sampling $B$ variational parameters are sampled uniformly (random), and in the KDE-based approach they are based on the learned density (KDE); and the parameters with the highest objective value are chosen as the solution. Since our primary focus is to devise methods that find high-quality parameters with a few quantum circuit evaluations, we use the learned optimization policy by RL to generate trajectories of length $B/2$, and we resume the trajectory from the best parameters found by using Nelder-Mead for the rest of $B/2$ evaluations. This approach is motivated by our observation that the RL-based approach reaches regions with high-quality solutions rather quickly, yet subsequent quantum circuit evaluations are spent without further improvements on the objective value \cite{khairy2019reinforcement}. The visualizations of the RL agent as it navigates the QAOA landscape for $p=1$ and the results of the pure RL-based method are reported in \cite{khairy2019reinforcement}. 

We compare our results with gradient-free methods for the following reasons. First, analytical gradients are in general not available, and evaluating gradients on a quantum computer is computationally expensive~\cite{crooks2018performance,guerreschi2017practical}. Under reasonable assumptions about the hardware~\cite{Guerreschi2019}, one evaluation of the objective takes $1$ second. Therefore, minimizing the number of objective evaluations needed for optimization is of utmost importance. Estimating gradient requires at least two evaluations for each parameter, making gradient-based methods noncompetitive within the budget of $192$ objective evaluations ($\approx200$ sec projected running time on a real quantum computer) that we chose for our benchmark. Second, gradients are sensitive to noise~\cite{zhu2019training}, which includes stochastic noise originating from the random nature of quantum computation, and the noise caused by hardware errors. This is typically addressed by increasing the number of samples used to estimate the objective. On the other hand, optimizers employed for hyperparameter optimization,  such as Bayesian optimization, require  a few hundred evaluations simply to initialize the surrogate model, ruling out these optimizers from our benchmark where the evaluation budget is $192$.

\begin{figure*}[ht]
\centering
\subfloat[$p=1$]{\includegraphics[width=0.7\columnwidth]{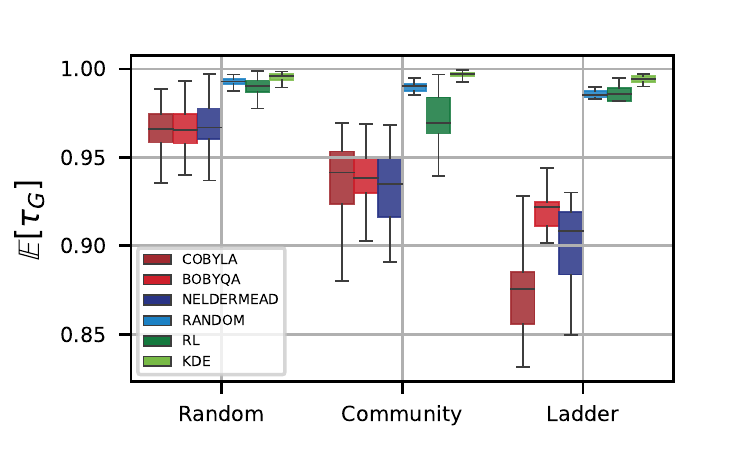}}
\hfill
\centering
\subfloat[$p=2$]{\includegraphics[width=0.7\columnwidth]{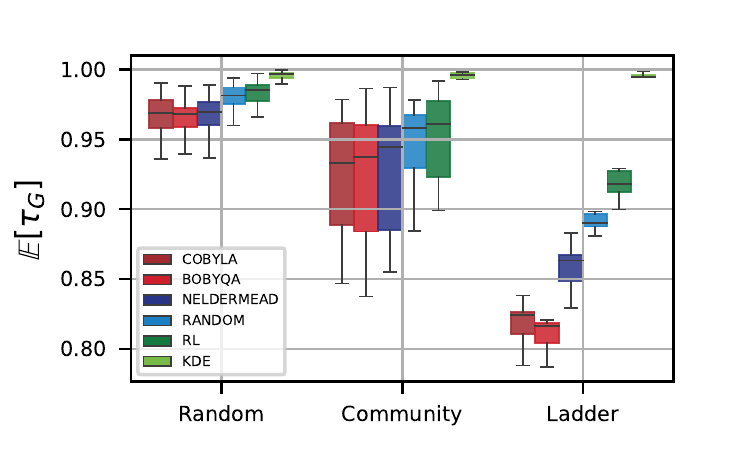}}
\hfill
\centering
\subfloat[$p=4$]{\includegraphics[width=0.7\columnwidth]{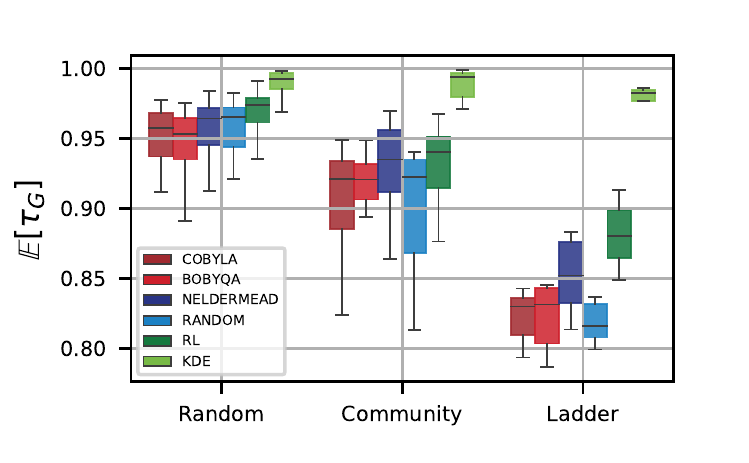}}
\hfill
\caption{Expected optimality ratio performance of different optimizers on graph instances in $G_\text{Test}$ for a given QAOA circuit depth $p \in \{1,2,4\}$. Test graph instances are grouped in three subgroups: random graphs, community graphs, and ladder graphs. Our proposed machine-learning-based methods, RL and KDE, outperform commonly used derivative-free off-the-shelf optimizers such as COBYLA, BOBYQA, and Nelder-Mead.}
\label{fig:optimizerPerformance}
\end{figure*}

To report the performance, we group graph instances in $G_\text{Test}$ in three subgroups, (1) Random graphs, which contains all graphs of the form $G_R(n_R,e_p)$, 2) Community graphs, which contains graphs of the form $G_C(n_C, n_k)$ and $G_B(n_B)$, and 3) Ladder graphs, which contains graphs of the form $G_L(n_L)$. In Figure \ref{fig:optimizerPerformance}, we report a boxplot of the expected optimality ratio, $\mathbb{E}[\tau_{G}] =  \mathbb{E}[f/f_\text{opt}]$, where the expectation is with respect to the highest objective value attained by a given optimizer in each of its $10$ attempts. The optimal solution to a graph instance in $G_\text{Test}$ is the largest known $f$ value found by any optimizer in any of its $10$ attempts for a given depth $p$. %Note that the quality of the optimal solution obtained by QAOA increases with depths $p$ (see Fig.~\ref{fig:QAOAPerformance}). 
One can see that the median optimality ratio achieved by our proposed approaches outperforms that of other commonly used optimizers. While the performance of derivative-free optimizers, random sampling, and the RL-based approach degrades as the dimension of parameters increase from $p=1$ to $p=4$, the KDE-based approach maintains high optimality ratios and is the superior approach in all cases and across different graph classes. The RL-based approach and random sampling rank second and third, respectively, in the majority of  cases. Random sampling turns out to be a competitive candidate for sampling of variational parameters, especially for low -epth $p=1$ circuits, and for random graph instances. We note that this finding hasd not been identified prior to our work because random search was not included in the experimental comparison \cite{shaydulin2019multistart,nakanishi2019sequential,larose2019variational,verdon2019learning}. %\PB{cite those works}.
In Table \ref{table3}, we summarize the median optimality gap reduction factor with respect to Nelder-Mead, attained by our proposed RL- and KDE-based approaches, namely, $(\frac{1-\mathbb{E}_\text{NM}[\tau_G]}{1-\mathbb{E}_\text{RL}[\tau_G]})$ and $(\frac{1-\mathbb{E}_\text{NM}[\tau_G]}{1-\mathbb{E}_\text{KDE}[\tau_G]})$. As the table shows,  our proposed approaches reduce the optimality gap by factors up to $30.15$ compared with Nelder-Mead, and the gap reduction factor is consistently larger than $1$.

\begin{table}[t] 
\caption{Median optimality gap reduction factor w.r.t Nelder-Mead.}\smallskip
\centering
\resizebox{.95\columnwidth}{!}{
\smallskip\begin{tabular}{|l|l|c|c|c|}
\hline
Graph Class & Proposed Optimizer & $p=1$ & $p=2$ & $p=4$ \\
\hline \hline
Random & RL & $3.07$ & $1.81$ & $1.37$ \\
 & KDE & $7.06$ & $8.07$ & $6.52$ \\
\hline \hline 
Community & RL & $2.66$ & $1.47$ & $1.05$ \\
 & KDE & $26.20$ & $14.15$ & $9.31$ \\
\hline \hline 
Ladder & RL & $5.59$ & $1.87$ & $1.12$ \\
 & KDE & $17.68$ & $30.15$ & $8.15$ \\
\hline 
\end{tabular} 
}
\label{table3}
\end{table}

\begin{figure}[ht]
\centering
\includegraphics[width=0.79\columnwidth]{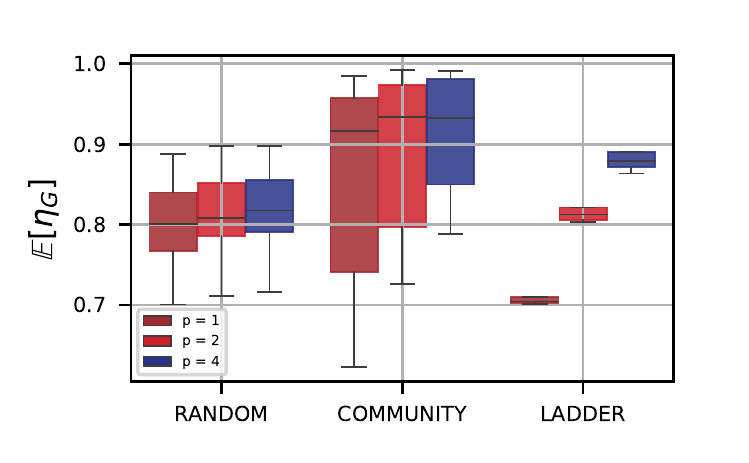}
\caption{Approximation ratio performance of QAOA with respect to classical optimal on $G_\text{Test}$. With higher-depth QAOA circuit, the attained approximation ratio increases.}
\label{fig:QAOAPerformance}
\end{figure}

In Figure \ref{fig:QAOAPerformance}, we show a boxplot of the expected approximation ratio performance, $\mathbb{E}[\eta_G] = \mathbb{E}[f/C_\text{opt}]$, of QAOA with respect to the classical optimal $C_\text{opt}$ found using brute force methods across different graph instances in $G_\text{Test}$. We can see that increasing the depth of QAOA circuit improves the attained approximation ratio, especially for structured graphs (i.e., community and ladder graph instances). 

\section{Conclusion}
\label{sec:con}

In this paper, we formulated the problem of finding optimal QAOA parameters for approximately solving combinatorial problems as a learning task. Two machine-learning-based approaches have been proposed: an RL-based approach, which can learn a policy network that can efficiently optimize new QAOA circuits by exploiting geometrical regularities in QAOA objectives, and a KDE-based approach, which can learn a generative model of optimal QAOA parameters that can be used to sample parameters for new QAOA circuits. Our proposed approaches have been trained on a small set of small-sized training instances, yet they are capable of efficiently solving larger problem instances. When coupled with QAOA, our proposed approaches can be powerful tools for amortizing the QAOA optimization cost across combinatorial instances. 

In our future work, we will investigate machine-learning-based methods for QAOA applied to constrained combinatorial optimization problems such as maximum independent set and max $\kappa$ colorable subgraphs, which have important applications in many disciplines. 

\subsection*{Acknowledgments}
This material is based upon work supported by the U.S. Department of Energy, Office of Science, Office of Advanced Scientific Computing Research, under Contract DE-AC02-06CH11357. This research was funded in part by and used resources of the Argonne Leadership Computing Facility, which is a DOE Office of Science User Facility supported under Contract DE-AC02-06CH11357. We gratefully acknowledge the computing resources provided on Bebop, a high-performance computing cluster operated by the Laboratory Computing Resource Center at Argonne National Laboratory. LC acknowledges support from LANL’s Laboratory Directed Research and Development (LDRD) program. LC was also supported by the U.S. Department of Energy, Office of Science, Office of Advanced Scientific Computing Research, under the Quantum Computing Application Teams program.

\bibliography{qaoa}

\begin{thebibliography}{}

\bibitem[\protect\citeauthoryear{Andrychowicz \bgroup et al\mbox.\egroup
  }{2016}]{andrychowicz2016learning}
Andrychowicz, M.; Denil, M.; Gomez, S.; Hoffman, M.~W.; Pfau, D.; Schaul, T.;
  Shillingford, B.; and De~Freitas, N.
\newblock 2016.
\newblock Learning to learn by gradient descent by gradient descent.
\newblock In {\em Advances in Neural Information Processing Systems},
  3981--3989.

\bibitem[\protect\citeauthoryear{Apolloni, Carvalho, and de
  Falco}{1989}]{Apolloni1989}
Apolloni, B.; Carvalho, C.; and de~Falco, D.
\newblock 1989.
\newblock Quantum stochastic optimization.
\newblock {\em Stochastic Processes and Their Applications} 33(2):233--244.

\bibitem[\protect\citeauthoryear{{Balaprakash} \bgroup et al\mbox.\egroup
  }{2018}]{8638041}
{Balaprakash}, P.; {Salim}, M.; {Uram}, T.; {Vishwanath}, V.; and {Wild}, S.
\newblock 2018.
\newblock {DeepHyper: Asynchronous} hyperparameter search for deep neural
  networks.
\newblock In {\em 2018 IEEE 25th International Conference on High Performance
  Computing (HiPC)},  42--51.

\bibitem[\protect\citeauthoryear{Balaprakash, Birattari, and
  St{\"u}tzle}{2007}]{prasanna2007}
Balaprakash, P.; Birattari, M.; and St{\"u}tzle, T.
\newblock 2007.
\newblock Improvement strategies for the {F-Race} algorithm: Sampling design
  and iterative refinement.
\newblock In et~al, B.-B., ed., {\em Hybrid Metaheuristics},  108--122.
\newblock Berlin, Heidelberg: Springer Berlin Heidelberg.

\bibitem[\protect\citeauthoryear{Barkoutsos \bgroup et al\mbox.\egroup
  }{2019}]{barkoutsos2019improving}
Barkoutsos, P.~K.; Nannicini, G.; Robert, A.; Tavernelli, I.; and Woerner, S.
\newblock 2019.
\newblock Improving variational quantum optimization using {CVaR}.
\newblock {\em arXiv preprint arXiv:1907.04769}.

\bibitem[\protect\citeauthoryear{Brandao \bgroup et al\mbox.\egroup
  }{2018}]{brandao2018fixed}
Brandao, F.~G.; Broughton, M.; Farhi, E.; Gutmann, S.; and Neven, H.
\newblock 2018.
\newblock For fixed control parameters the quantum approximate optimization
  algorithm's objective function value concentrates for typical instances.
\newblock {\em arXiv:1812.04170}.

\bibitem[\protect\citeauthoryear{Crooks}{2018}]{crooks2018performance}
Crooks, G.~E.
\newblock 2018.
\newblock Performance of the quantum approximate optimization algorithm on the
  maximum cut problem.
\newblock {\em arXiv:1811.08419}.

\bibitem[\protect\citeauthoryear{Farhi, Goldstone, and
  Gutmann}{2014}]{farhi2014quantum}
Farhi, E.; Goldstone, J.; and Gutmann, S.
\newblock 2014.
\newblock A quantum approximate optimization algorithm.
\newblock {\em arXiv:1411.4028}.

\bibitem[\protect\citeauthoryear{Feurer, Springenberg, and
  Hutter}{2015}]{feurer2015initializing}
Feurer, M.; Springenberg, J.~T.; and Hutter, F.
\newblock 2015.
\newblock Initializing {B}ayesian hyperparameter optimization via
  meta-learning.
\newblock In {\em Twenty-Ninth AAAI Conference on Artificial Intelligence}.

\bibitem[\protect\citeauthoryear{Goemans and
  Williamson}{1995}]{goemans1995improved}
Goemans, M.~X., and Williamson, D.~P.
\newblock 1995.
\newblock Improved approximation algorithms for maximum cut and satisfiability
  problems using semidefinite programming.
\newblock {\em Journal of the ACM} 42(6):1115--1145.

\bibitem[\protect\citeauthoryear{Guerreschi and
  Matsuura}{2019}]{Guerreschi2019}
Guerreschi, G.~G., and Matsuura, A.~Y.
\newblock 2019.
\newblock {QAOA} for max-cut requires hundreds of qubits for quantum speed-up.
\newblock {\em Scientific Reports} 9(1).

\bibitem[\protect\citeauthoryear{Guerreschi and
  Smelyanskiy}{2017}]{guerreschi2017practical}
Guerreschi, G.~G., and Smelyanskiy, M.
\newblock 2017.
\newblock Practical optimization for hybrid quantum-classical algorithms.
\newblock {\em arXiv:1701.01450}.

\bibitem[\protect\citeauthoryear{Hadfield \bgroup et al\mbox.\egroup
  }{2017}]{hadfield2017quantum}
Hadfield, S.; Wang, Z.; O'Gorman, B.; Rieffel, E.~G.; Venturelli, D.; and
  Biswas, R.
\newblock 2017.
\newblock From the quantum approximate optimization algorithm to a quantum
  alternating operator ansatz.
\newblock {\em arXiv preprint arXiv:1709.03489}.

\bibitem[\protect\citeauthoryear{Huang \bgroup et al\mbox.\egroup
  }{2019}]{huang2019alibaba}
Huang, C.; Szegedy, M.; Zhang, F.; Gao, X.; Chen, J.; and Shi, Y.
\newblock 2019.
\newblock Alibaba cloud quantum development platform: Applications to quantum
  algorithm design.
\newblock {\em arXiv preprint arXiv:1909.02559}.

\bibitem[\protect\citeauthoryear{Hutter, Kotthoff, and
  Vanschoren}{2018}]{automl_book}
Hutter, F.; Kotthoff, L.; and Vanschoren, J., eds.
\newblock 2018.
\newblock {\em Automated Machine Learning: Methods, Systems, Challenges}.
\newblock Springer.
\newblock available at http://automl.org/book.

\bibitem[\protect\citeauthoryear{Johnson}{2019}]{nlopt}
Johnson, S.~G.
\newblock 2019.
\newblock The {NLopt} nonlinear-optimization package.

\bibitem[\protect\citeauthoryear{Kato}{1950}]{Kato1950}
Kato, T.
\newblock 1950.
\newblock On the adiabatic theorem of quantum mechanics.
\newblock {\em Journal of the Physical Society of Japan} 5(6):435--439.

\bibitem[\protect\citeauthoryear{Khairy \bgroup et al\mbox.\egroup
  }{2019}]{khairy2019reinforcement}
Khairy, S.; Shaydulin, R.; Cincio, L.; Alexeev, Y.; and Balaprakash, P.
\newblock 2019.
\newblock Reinforcement-learning-based variational quantum circuits
  optimization for combinatorial problems.
\newblock {\em arXiv preprint arXiv:1911.04574}.

\bibitem[\protect\citeauthoryear{LaRose \bgroup et al\mbox.\egroup
  }{2019}]{larose2019variational}
LaRose, R.; Tikku, A.; O’Neel-Judy, {\'E}.; Cincio, L.; and Coles, P.~J.
\newblock 2019.
\newblock Variational quantum state diagonalization.
\newblock {\em npj Quantum Information} 5(1):8.

\bibitem[\protect\citeauthoryear{Li and Malik}{2016}]{li2016learning}
Li, K., and Malik, J.
\newblock 2016.
\newblock Learning to optimize.
\newblock {\em arXiv preprint arXiv:1606.01885}.

\bibitem[\protect\citeauthoryear{Nakanishi, Fujii, and
  Todo}{2019}]{nakanishi2019sequential}
Nakanishi, K.~M.; Fujii, K.; and Todo, S.
\newblock 2019.
\newblock Sequential minimal optimization for quantum-classical hybrid
  algorithms.
\newblock {\em arXiv:1903.12166}.

\bibitem[\protect\citeauthoryear{Pagano \bgroup et al\mbox.\egroup
  }{2019}]{pagano2019quantum}
Pagano, G.; Bapat, A.; Becker, P.; Collins, K.; De, A.; Hess, P.; Kaplan, H.;
  Kyprianidis, A.; Tan, W.; Baldwin, C.; et~al.
\newblock 2019.
\newblock Quantum approximate optimization with a trapped-ion quantum
  simulator.
\newblock {\em arXiv preprint arXiv:1906.02700}.

\bibitem[\protect\citeauthoryear{Parekh, Ryan-Anderson, and
  Gharibian}{2019}]{osti_1492737}
Parekh, O.~D.; Ryan-Anderson, C.; and Gharibian, S.
\newblock 2019.
\newblock Quantum optimization and approximation algorithms.
\newblock Technical report.

\bibitem[\protect\citeauthoryear{Romero \bgroup et al\mbox.\egroup
  }{2018}]{romero2018strategies}
Romero, J.; Babbush, R.; McClean, J.; Hempel, C.; Love, P.; and Aspuru-Guzik,
  A.
\newblock 2018.
\newblock Strategies for quantum computing molecular energies using the unitary
  coupled cluster ansatz.
\newblock {\em Quantum Science and Technology}.

\bibitem[\protect\citeauthoryear{Schulman \bgroup et al\mbox.\egroup
  }{2017}]{schulman2017proximal}
Schulman, J.; Wolski, F.; Dhariwal, P.; Radford, A.; and Klimov, O.
\newblock 2017.
\newblock Proximal policy optimization algorithms.
\newblock {\em arXiv preprint arXiv:1707.06347}.

\bibitem[\protect\citeauthoryear{Shaydulin \bgroup et al\mbox.\egroup
  }{2018}]{shaydulin2018community}
Shaydulin, R.; Ushijima-Mwesigwa, H.; Safro, I.; Mniszewski, S.; and Alexeev,
  Y.
\newblock 2018.
\newblock Community detection across emerging quantum architectures.
\newblock {\em Proceedings of the 3rd International Workshop on Post Moore's
  Era Supercomputing}.

\bibitem[\protect\citeauthoryear{Shaydulin \bgroup et al\mbox.\egroup
  }{2019}]{shaydulin2018network}
Shaydulin, R.; Ushijima-Mwesigwa, H.; Safro, I.; Mniszewski, S.; and Alexeev,
  Y.
\newblock 2019.
\newblock Network community detection on small quantum computers.
\newblock {\em Advanced Quantum Technologies}  1900029.

\bibitem[\protect\citeauthoryear{Shaydulin, Safro, and
  Larson}{2019}]{shaydulin2019multistart}
Shaydulin, R.; Safro, I.; and Larson, J.
\newblock 2019.
\newblock Multistart methods for quantum approximate optimization.
\newblock {\em 2019 {IEEE} High Performance Extreme Computing Conference
  ({HPEC})}.

\bibitem[\protect\citeauthoryear{Streif and Leib}{2019}]{streif2019comparison}
Streif, M., and Leib, M.
\newblock 2019.
\newblock Comparison of {QAOA} with quantum and simulated annealing.
\newblock {\em arXiv preprint arXiv:1901.01903}.

\bibitem[\protect\citeauthoryear{Sutton and
  Barto}{2018}]{sutton2018reinforcement}
Sutton, R.~S., and Barto, A.~G.
\newblock 2018.
\newblock {\em Reinforcement learning: An introduction}.
\newblock MIT Press.

\bibitem[\protect\citeauthoryear{Verdon \bgroup et al\mbox.\egroup
  }{2019}]{verdon2019learning}
Verdon, G.; Broughton, M.; McClean, J.~R.; Sung, K.~J.; Babbush, R.; Jiang, Z.;
  Neven, H.; and Mohseni, M.
\newblock 2019.
\newblock Learning to learn with quantum neural networks via classical neural
  networks.
\newblock {\em arXiv preprint arXiv:1907.05415}.

\bibitem[\protect\citeauthoryear{Wang \bgroup et al\mbox.\egroup
  }{2018}]{PhysRevA.97.022304}
Wang, Z.; Hadfield, S.; Jiang, Z.; and Rieffel, E.~G.
\newblock 2018.
\newblock Quantum approximate optimization algorithm for {MaxCut:} a fermionic
  view.
\newblock {\em Physical Review A} 97:022304.

\bibitem[\protect\citeauthoryear{Wecker, Hastings, and
  Troyer}{2016}]{wecker2016training}
Wecker, D.; Hastings, M.~B.; and Troyer, M.
\newblock 2016.
\newblock Training a quantum optimizer.
\newblock {\em Physical Review A} 94(2):022309.

\bibitem[\protect\citeauthoryear{Yang \bgroup et al\mbox.\egroup
  }{2017}]{yang2017optimizing}
Yang, Z.-C.; Rahmani, A.; Shabani, A.; Neven, H.; and Chamon, C.
\newblock 2017.
\newblock Optimizing variational quantum algorithms using {P}ontryagin's
  minimum principle.
\newblock {\em Physical Review X} 7(2):021027.

\bibitem[\protect\citeauthoryear{Zhou \bgroup et al\mbox.\egroup
  }{2018}]{zhou2018quantum}
Zhou, L.; Wang, S.-T.; Choi, S.; Pichler, H.; and Lukin, M.~D.
\newblock 2018.
\newblock Quantum approximate optimization algorithm: Performance, mechanism,
  and implementation on near-term devices.
\newblock {\em arXiv:1812.01041}.

\bibitem[\protect\citeauthoryear{Zhu \bgroup et al\mbox.\egroup
  }{2019}]{zhu2019training}
Zhu, D.; Linke, N.~M.; Benedetti, M.; Landsman, K.~A.; Nguyen, N.~H.; Alderete,
  C.~H.; Perdomo-Ortiz, A.; Korda, N.; Garfoot, A.; Brecque, C.; et~al.
\newblock 2019.
\newblock Training of quantum circuits on a hybrid quantum computer.
\newblock {\em Science advances} 5(10):eaaw9918.

\end{thebibliography}
\bibliographystyle{aaai}

\begin{center}
    \framebox{\parbox{3in}{
    The submitted manuscript has been created by UChicago Argonne, LLC, Operator of Argonne National Laboratory (``Argonne''). Argonne, a U.S. Department of Energy Office of Science laboratory, is operated under Contract No. DE-AC02-06CH11357. The U.S. Government retains for itself, and others acting on its behalf, a paid-up nonexclusive, irrevocable worldwide license in said article to reproduce, prepare derivative works, distribute copies to the public, and perform publicly and display publicly, by or on behalf of the Government. The Department of Energy will provide public access to these results of federally sponsored research in accordance with the DOE Public Access Plan. \url{http://energy.gov/downloads/doe-public-access-plan}}}
    \normalsize
\end{center}

\end{document}